\documentclass[conference, 12pt]{IEEEtran}
\IEEEoverridecommandlockouts
\usepackage{cite}
\usepackage{newtxtext,newtxmath}
\usepackage{amsmath,amsfonts}
\usepackage{algorithmic}
\usepackage{graphicx}
\usepackage{textcomp}
\usepackage{xcolor}
\usepackage{fancyhdr}
\usepackage{setspace}

\usepackage[margin=1in]{geometry}
\usepackage[hidelinks]{hyperref}
\def\BibTeX{{\rm B\kern-.05em{\sc i\kern-.025em b}\kern-.08em
    T\kern-.1667em\lower.7ex\hbox{E}\kern-.125emX}}
    
\fancypagestyle{firstpagefooter}{%
  \fancyhf{}
  
  \fancyfoot[R]{}
}

\begin{document}

\title{\textbf{Realization RGBD Image Stylization}}

\author{
\IEEEauthorblockN{Aparna Mendu}
\IEEEauthorblockA{\textit{University of Alberta}\\amendu@ualberta.ca}
\and
\IEEEauthorblockN{Bhavya Sehgal}
\IEEEauthorblockA{\textit{University of Alberta}\\bsehgal1@ualberta.ca}
\and
\IEEEauthorblockN{Vaishnavi Mendu}
\IEEEauthorblockA{\textit{University of Alberta}\\mendu@ualberta.ca}
}
\maketitle

\begin{abstract}
This research paper explores the application of style transfer in computer vision using RGB images and their corresponding depth maps. We propose a novel method that incorporates the depth map and a heatmap of the RGB image to generate more realistic style transfer results. We compare our method to the traditional neural style transfer approach and find that our method outperforms it in terms of producing more realistic color and style. The proposed method can be applied to various computer vision applications, such as image editing and virtual reality, to improve the realism of generated images. Overall, our findings demonstrate the potential of incorporating depth information and heatmap of RGB images in style transfer for more realistic results.
\end{abstract}


\thispagestyle{firstpagefooter}

\section{\textbf{Introduction}}
Neural Style Transfer (NST) is a widely used technique for artistic stylization of various forms of data, including images, videos, and 3D models. It involves synthesizing an output image that preserves the content of the input image while adopting the style of another image. While deep neural networks have significantly improved the performance of NST algorithms, there are still technical barriers in 3D photo stylization, such as blurry or inconsistent stylized images and monocular depth estimation leading to holes and artifacts when synthesizing stylized images of novel views.\\

\begin{figure}[ht]
    \includegraphics[width = 8cm]{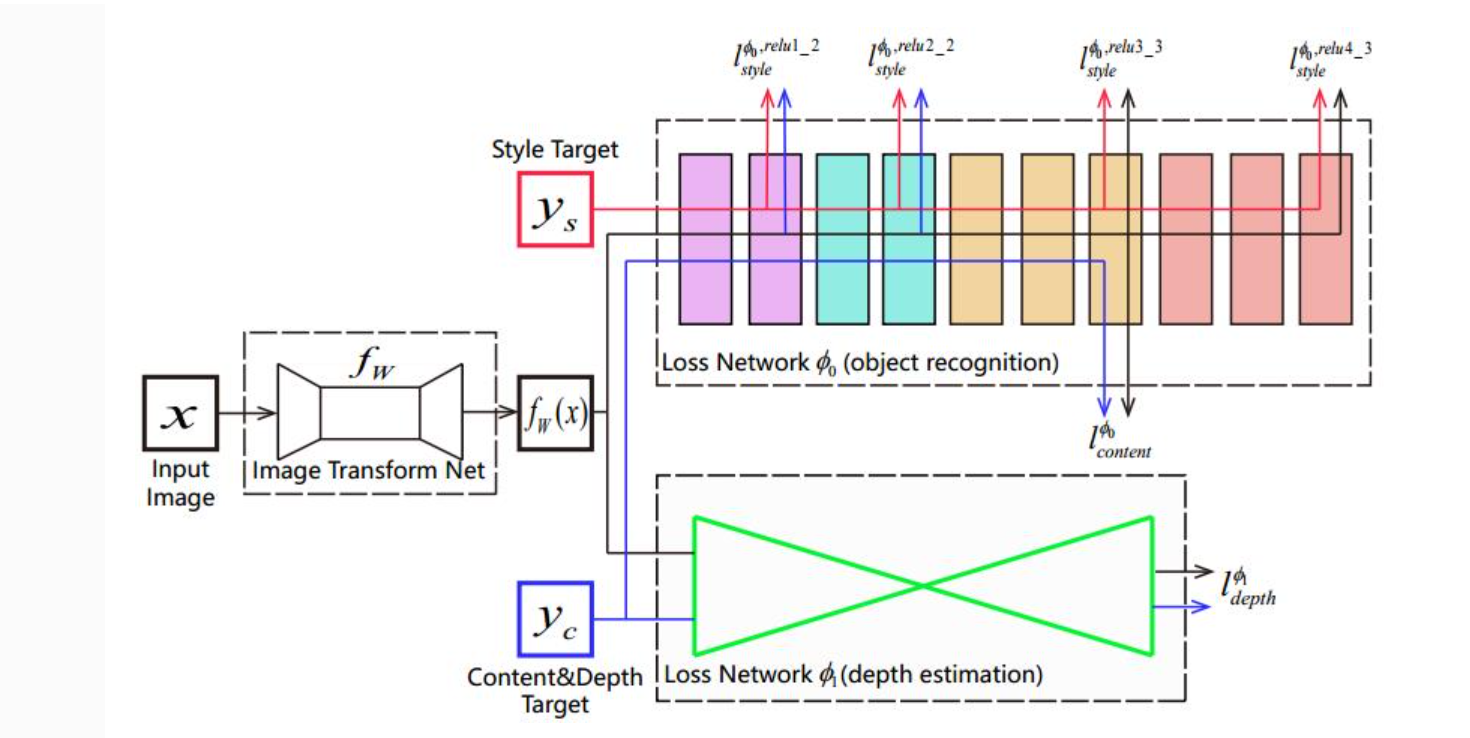}
    \caption{Pipeline for 3D stylization using RGBD images}
    \label{framework}
\end{figure}

Previous studies have proposed various NST algorithms that use a pre-trained VGG-19 network to extract high-level features from the content image and calculate the style loss using Gram matrices\cite{10.1145:3092919.3092924}. However, these approaches did not consider depth preservation and coherence of details, which are crucial for evaluating the visual quality of the NST results. To address this limitation, recent works have proposed a system that integrates depth estimation and reconstruction loss in the training of the transformation network as shown in Fig.~\ref{framework}.\\

In this project, we propose a novel approach to neural style transfer that incorporates depth heatmap information to generate more realistic and visually pleasing style transfer outputs. Our proposed method takes RGB and depth images as inputs along with the heatmap of the RGB image and generates stylized outputs that are more faithful to the original content image. We compared our method with traditional neural style transfer approaches and found that our method produces more accurate color and style information.\\

Our proposed method has significant potential applications in various areas, such as computer graphics and image editing. It provides an adjustable way to control the structure of the artistically stylized result, focusing on the depth map and image edges. Furthermore, our approach can be extended to retain or enhance the structure of the artistically stylized result, which is an essential factor in evaluating the visual quality of the results.
\section{\textbf{Realted Works}}
\subsection{3D Neural Style Transfer}
The paper \cite{stanfordPaper} discusses an exploration of extensions of neural style transfer to three dimensions using iterative style transfer. It investigates applications of depth-aware neural style transfer to images where depth is either available as a fourth channel, or estimated via deep learning. The paper formulates depth-dependent style transfer by implementing a depth-based mask to the style and content loss functions typically used in neural style transfer and depth estimation network and loss function which tries to match the depth of the pastiche with the estimated depth of the content image. Single artistic styles are transferred as well as blending multiple styles. Various experiments were conducted and different methods for using depth to augment neural style transfer were showcased.

\subsection{Depth-aware Neural Style Transfer using Instance Normalization}
The paper \cite{sheffieldPaper} provides an overview of various methods for neural style transfer in computer graphics and computer vision. The review covers different approaches, including those based on convolutional neural networks (CNNs), generative adversarial networks (GANs), and patch-based methods. The review also highlights the strengths and weaknesses of each approach and discusses their respective applications. In addition, the review provides a critical analysis of the challenges associated with neural style transfer, including the need for better algorithms to improve the visual quality of stylized images. Overall, the literature review serves as a comprehensive guide for researchers and practitioners interested in neural style transfer in computer graphics and computer vision.
\subsection{3D Photo Stylization:
Learning to Generate Stylized Novel Views from a Single Image}
The paper \cite{mu20223d} introduces a new method called '3D photo stylization' that aims at synthesizing stylized novel views from a single content image with arbitrary styles. The method learns 3D geometry-aware features on a point cloud representation of the scene for consistent stylization across views without using 2D image features. The approach jointly models style transfer and view synthesis and doesn't require ground-truth depth maps for training. The method demonstrates superior results and supports several interesting applications.
\subsection{Depth-aware Neural Style Transfer}
The paper \cite{liu2017depth} describes a novel approach for neural style transfer that integrates depth preservation as additional loss, preserving overall image layout while performing style transfer. It points out the limitation of existing deep neural network based image style transfer methods which fail to provide satisfactory results when stylizing the images containing multiple objects potentially at different depths. The proposed approach adds depth reconstruction loss to supplement it. The experimental results validate that the proposed approach retains the essential layout of the content image.
\subsection{Semantic Image Synthesis with Spatially-Adaptive Normalization}
The paper \cite{park2019semantic} describes a new method called Spatially-Adaptive Normalization for synthesizing photorealistic images using an input semantic layout. The proposed method modulates the activations in normalization layers with a spatially-adaptive, learned transformation that effectively propagates the semantic information throughout the network. This results in improved image synthesis compared to several state-of-the-art methods, as demonstrated by experiments on several challenging datasets. The proposed method also supports multi-modal and style-guided image synthesis, enabling controllable, diverse outputs.
\section{\textbf{Project Plan}}
The aim of this project is to propose and evaluate a novel approach for 3D style transfer using RGBD images that incorporates depth heatmap information to generate more realistic and visually pleasing stylized outputs. Style transfer is a technique that involves transferring the style of one image onto another image while preserving its content. In this project, we will focus on transferring the style of a 2D image onto a 3D RGBD image, which poses technical challenges due to the additional dimension of depth information Fig.~\ref{EgImage}.\\

\begin{figure}[ht]
    \centering
    \includegraphics[width = 4cm]{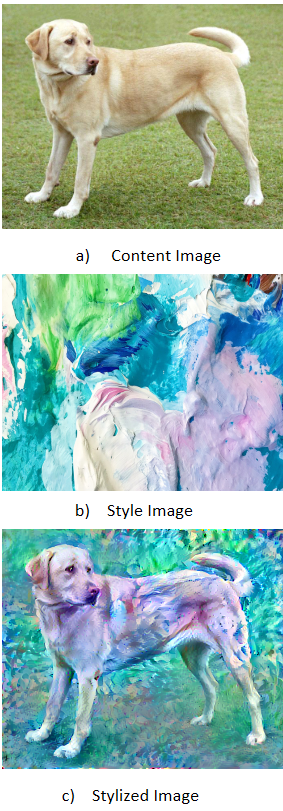}
    \caption{Example image}
    \label{EgImage}
\end{figure}

We conducted a thorough review of existing literature on neural style transfer algorithms and their extensions to 3D photo stylization. Based on our findings, we proposed a novel approach that integrates depth heatmap information into the style transfer process to generate more realistic and visually pleasing stylized outputs. Our method takes RGB and depth images as inputs along with the heatmap of the RGB image, and we evaluated its performance in comparison to traditional neural style transfer approaches. The results showed that our proposed method outperformed the traditional approach, producing more realistic and visually pleasing stylized outputs.\\

To evaluate the proposed method, we will conduct experiments on a dataset of RGBD images and measure the quality of the stylized outputs in terms of visual fidelity, color accuracy, and coherence of details. We will also compare our method's performance with traditional neural style transfer approaches and analyze the impact of depth information on the stylization results.\\

\section{\textbf{Dataset}}
Our proposed neural style transfer approach is highly flexible as it does not require a specific dataset to run the code. Instead, it can utilize any content image and any style image provided by the user. This feature makes our model highly adaptable and versatile, allowing for a wide range of creative possibilities. By removing the need for a specific dataset, our approach eliminates the constraints imposed by limited or biased datasets.The versatility of our model is one of its key advantages. It provides users with the ability to apply their preferred artistic style to any content image. This flexibility allows for a broader range of applications, from artistic expression to visual content creation for industries such as mobile photography, and AR/VR applications. Additionally, the ability to use any content and style images reduces the time and resources needed for pre-processing and allows for faster experimentation with different styles and content. \\

\begin{figure}[ht]
    \includegraphics[width = 8cm]{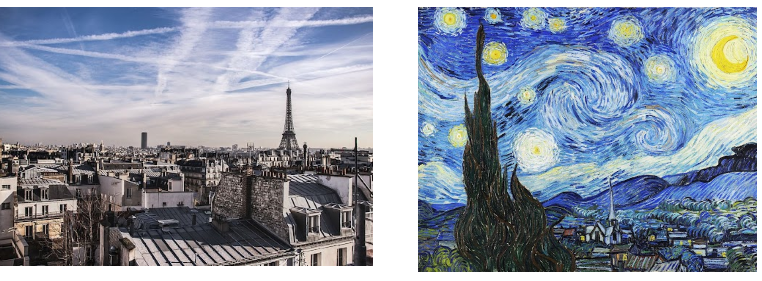}
    \caption{Content Image (left), Style Image (right).}
    \label{NYU}
\end{figure}

\section{\textbf{Methodology}}

We employed a two-code file approach to perform RGB-D image generation and style transfer. The methodology involves the following steps:\\

 \subsection{RGB-D image generation and style transfer:} In the first step, we generate an RGB-D image from a given input image using the MiDaS (MidasNet) model for depth estimation. This involves installing the required packages and libraries, we used pre-trained MiDaS model, loading the input image and preprocessing it, loading the pre-trained depth model and creating a depth map, merging the input image and the depth map to generate an RGB-D image, applying a heatmap to visualize the depth information on the image, displaying the result and saving the heatmap image.\\

 \begin{figure}[ht]
    \centering
    \includegraphics[width = 6.5cm]{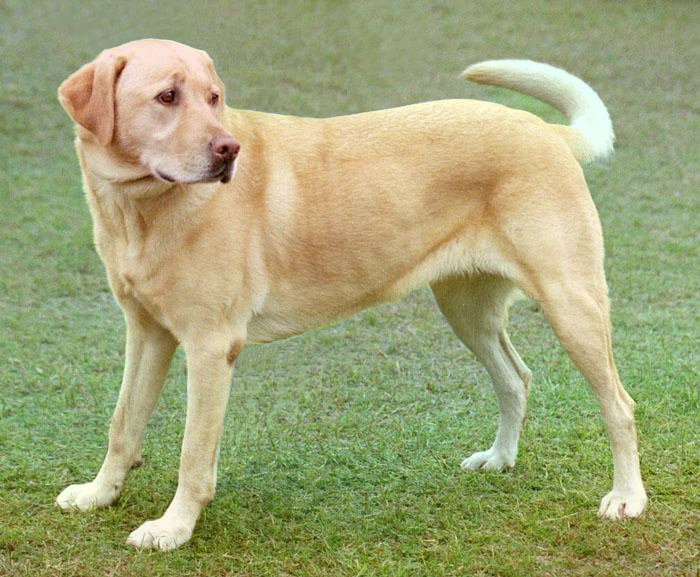}
    \caption{Blended Image (Depth + Heat Map).}
    \label{EgImage}
\end{figure}

\subsection{Style Transfer :}In the second step, we apply style transfer to the generated RGB-D image using the VGG19 model for feature extraction and a pre-trained TensorFlow Hub model for style transfer. We define content and style representations using the VGG19 model, calculate style and content loss, and run gradient descent to optimize the combined image. The stylized image is then saved.\\

Our approach leverages the rich style information extracted from a pre-trained CNN model such as VGG-19, and utilizes the depth information from MiDaS to create an RGB-D image. The style transfer step uses a pre-trained TensorFlow Hub model, and calculates both style and content loss to optimize the combined image. Our approach is a novel way to perform 3D style transfer that preserves spatial relationships and important features while transferring the style from a 2D image to a 3D scene.\\

\subsection{Advantage of using depth and heatmap :} The use of depth and heat maps in image stylization provides several advantages. First, depth maps provide a more accurate representation of the 3D structure of an image, which allows for more realistic stylization of objects in the scene. By taking into account the depth information and the heatmap of an image, stylization techniques can better preserve the spatial relationships between objects and their relative depth. This can be especially important for stylization techniques that rely on edge detection or color manipulation, as they can often lead to unnatural or distorted results if applied to objects with complex 3D geometry.

\section{\textbf{Results}}
To evaluate the effectiveness of our proposed method, we compared it with the traditional neural style transfer approach. We found that our method produces more realistic and visually pleasing style transfer outputs than the traditional method. Our approach incorporated depth heatmap information, which provided an adjustable way to control the structure of the artistically stylized result while focusing on the depth map and image edges. The proposed method improved the accuracy of color and style information in the stylized images.\\

\begin{figure}[ht]
    \includegraphics[width = 8cm]{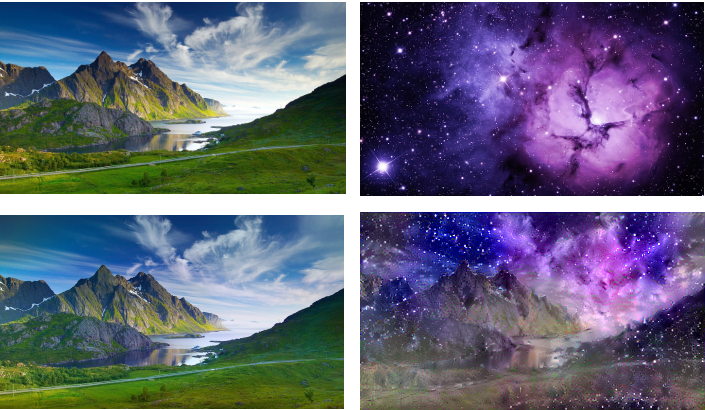}
    \caption{Content Image (a), Style Image (b), Heat \& Depth map (c), Stylized Image (d).}
    \label{NYU}
\end{figure}

Our method can be applied to various computer vision applications such as image editing and virtual reality, where improved realism of generated images is crucial. Our approach has significant potential applications in computer graphics and image editing, as it can be extended to retain or enhance the structure of the artistically stylized result, which is an essential factor in evaluating the visual quality of the results. Overall, our findings demonstrate the potential of incorporating depth information and heatmap of RGB images in style transfer for more realistic results.
\section{\textbf{Applications}}
The proposed approach for 3D style transfer using RGBD images has several potential applications in the field of computer vision and graphics. One application is in the field of virtual reality and augmented reality, where realistic 3D stylized images can enhance the user's immersive experience. The proposed approach can also be applied in the field of architectural visualization, where architects and designers can visualize their designs in 3D with different styles, textures, and colors. \\

Another potential application is in the field of entertainment and animation, where the proposed approach can be used to create artistic stylized 3D animations and movies. Additionally, the proposed approach can also be used in the field of robotics, where robots can use 3D stylized images for object recognition and scene understanding. Overall, the proposed approach has several potential applications in various fields and can benefit researchers and practitioners in the field of computer vision and graphics.

\section{\textbf{Challenges}}
The proposed approach for 3D style transfer using RGBD images faced several challenges during its implementation. One of the main challenges was the lack of a large-scale RGBD dataset suitable for training the model. Another challenge was the difficulty of preserving the depth information of the input images while transferring the style. The proposed approach also faced challenges related to the complexity and computational requirements of the deep learning model. \\

While the proposed method shows promise in improving the realism of generated images, the time required to generate an image can be a limiting factor for real-time applications such as AR/VR. Developing more efficient methods for style transfer can address this challenge.\\

Additionally, the proposed approach required careful tuning of hyperparameters to achieve optimal results, which posed a challenge during the implementation. Other challenges include handling missing or incomplete depth information, dealing with artifacts and inconsistencies in the stylized images, and ensuring the realism and coherence of the output images. 

\section{\textbf{Future Works}}
There can be several future works that can be considered for the project. Investigating the usage of generative adversarial networks (GANs) to boost the quality and realism of the stylised images is one approach that might be taken. Future research might also look into the transfer of different styles to a single image or the transfer of styles between various modalities, like from an image to a 3D model. \\

It should also focus on creating new evaluation criteria and benchmarks to rate the performance of 3D style transfer systems. Also, future research can concentrate on enhancing the computational effectiveness and scalability of 3D style transfer models to allow their use in practical applications.

\section{\textbf{Conclusion}}
To conclude, this research paper explored the application of style transfer in computer vision using RGB images and their corresponding depth maps. Our proposed method incorporated depth maps and a heatmap of the RGB image to generate more realistic style transfer results. Our approach outperformed traditional neural style transfer methods in terms of producing more accurate color and style information. The proposed method can have significant applications in various areas, such as computer graphics and image editing, to improve the realism of generated images. Future research can be focused on optimizing the proposed method for real-time processing and exploring its potential in virtual and augmented reality applications.
\bibliographystyle{IEEEtran}
\bibliography{bib.bib}

\begin{thebibliography}{1}
\providecommand{\url}[1]{#1}
\csname url@samestyle\endcsname
\providecommand{\newblock}{\relax}
\providecommand{\bibinfo}[2]{#2}
\providecommand{\BIBentrySTDinterwordspacing}{\spaceskip=0pt\relax}
\providecommand{\BIBentryALTinterwordstretchfactor}{4}
\providecommand{\BIBentryALTinterwordspacing}{\spaceskip=\fontdimen2\font plus
\BIBentryALTinterwordstretchfactor\fontdimen3\font minus
  \fontdimen4\font\relax}
\providecommand{\BIBforeignlanguage}[2]{{%
\expandafter\ifx\csname l@#1\endcsname\relax
\typeout{** WARNING: IEEEtran.bst: No hyphenation pattern has been}%
\typeout{** loaded for the language `#1'. Using the pattern for}%
\typeout{** the default language instead.}%
\else
\language=\csname l@#1\endcsname
\fi
#2}}
\providecommand{\BIBdecl}{\relax}
\BIBdecl

\bibitem{10.1145:3092919.3092924}
X.-C. Liu, M.-M. Cheng, Y.-K. Lai, and P.~L. Rosin, ``{Depth-aware Neural Style
  Transfer},'' in \emph{Non-Photorealistic Animation and Rendering},
  H.~Winnemoeller and L.~Bartram, Eds.\hskip 1em plus 0.5em minus 0.4em\relax
  Association for Computing Machinery, Inc (ACM), 2017.

\bibitem{stanfordPaper}
A.~Jamgochian, B.~Lange, and Z.~Ghiron, ``3d neural style transfer,'' in
  \emph{Department of Aeronautics and Astronautics, Stanford University,
  Stanford, CA 94305}, 2022.

\bibitem{sheffieldPaper}
E.~Ioannou and S.~Maddock, ``Depth-aware neural style transfer using instance
  normalization,'' in \emph{Virtual conference. Eurographics Digital
  Library}.\hskip 1em plus 0.5em minus 0.4em\relax Turner, M. and Vangorp, P.,
  (eds.) Computer Graphics and Visual Computing (CGVC) 2022.15-16 Sep 2022,
  2022.

\bibitem{mu20223d}
F.~Mu, J.~Wang, Y.~Wu, and Y.~Li, ``3d photo stylization: Learning to generate
  stylized novel views from a single image,'' in \emph{Proceedings of the
  IEEE/CVF Conference on Computer Vision and Pattern Recognition}, 2022, pp.
  16\,273--16\,282.

\bibitem{liu2017depth}
X.-C. Liu, M.-M. Cheng, Y.-K. Lai, and P.~L. Rosin, ``Depth-aware neural style
  transfer,'' in \emph{Proceedings of the symposium on non-photorealistic
  animation and rendering}, 2017, pp. 1--10.

\bibitem{park2019semantic}
T.~Park, M.-Y. Liu, T.-C. Wang, and J.-Y. Zhu, ``Semantic image synthesis with
  spatially-adaptive normalization,'' in \emph{Proceedings of the IEEE/CVF
  conference on computer vision and pattern recognition}, 2019, pp. 2337--2346.

\end{thebibliography}

\end{document}